\documentclass[runningheads]{llncs}
\usepackage{graphicx}
\graphicspath{{images/}}
\usepackage{comment}

\begin{document}
\title{Ensemble Learning techniques for object detection in high-resolution satellite images}
\author{Arthur Vilhelm \and
Matthieu Limbert \and
Clement Audebert \and
Tugdual Ceillier}

\authorrunning{A. Vilhelm et al.}
\titlerunning{Ensemble Learning techniques for object detection}% Part of RIGHT running header
\institute{Earthcube, 75009 Paris, France \\ \url{www.earthcube.eu}}
\maketitle              % typeset the header of the contribution

\begin{abstract}
Ensembling is a method that aims to maximize the detection performance by fusing individual detectors. While rarely mentioned in deep-learning articles applied to remote sensing, ensembling methods have been widely used to achieve high scores in recent data science competitions, such as Kaggle. The few remote sensing articles mentioning ensembling mainly focus on mid resolution images and earth observation applications such as land use classification, but never on Very High Resolution (VHR) images for defense-related applications or object detection. This study aims at reviewing the most relevant ensembling techniques to be used for object detection on very high resolution imagery and shows an example of the value of such technique on a relevant operational use case (vehicle detection in desert areas).

\end{abstract}
%
%
%%%%%%%%%%%%%%%%%%%%%% Intro
\section{Introduction}
\subsection{Object detection}
Recent increases in the quality and volume of Very High Resolution (VHR) Earth observation images from commercially-available satellites has allowed breakthroughs in tools that can be used for mapping and processing objects and areas of interest to the defence and security communities.

Image analysts can only analyse a small part of all this data and may take hours to find a vehicle in a desert area. Computer vision algorithms and deep learning solutions applied to satellite images allow the development of automatic tools to monitor large areas in strategic regions. Large quantities of commercial satellite imagery of very high quality can be obtained to label and train algorithms to detect, classify and identify infrastructures such as aircrafts, vehicles, vessels, roads or buildings. At a time when it has become easy to classify cats and dogs in various situations, earth observation raises new challenges by the diversity of the environment types and conditions, and by the expectations of end-users.

Ensembling is an approach using a combination of multiple algorithms to achieve better performances than could be obtained from any of the constituent machine learning algorithm alone. Several academic articles describe the theoretical basis of ensemble learning with first implementations starting in the 1990's. While rarely mentioned in deep-learning papers applied to remote sensing, ensembling methods have been widely used to achieve high scores in recent Kaggle competitions. Remote sensing papers mentioning ensembling mainly focus on mid resolution images and earth observation applications such as land use classification, but never on VHR images for high resolution segmentation.

\subsection{Ensemble of neural networks}
Convolutional neural networks (CNN) is the most promising machine learning algorithm for object detection and classification in images. It is a class of supervised algorithms that takes advantage of a large dataset of labeled objects to learn a generalized model of this data. At prediction time, the association of convolutional kernels with efficient GPU make CNN an efficient tool with the potential for human-level accuracy.

Ensemble techniques are used in a wide variety of problems of machine learning. By combining several algorithms, it is known to reduce the bias of a single detector or its variance. Those problems of bias and variance of prediction are well known and can be caused by using a model that is too small or too large compared to the dataset or by the fact that earth observation data is too heterogeneous for the model to capture the full picture. In other words, there is not a single model that can properly model the full set of data to process. For instance, stacking is known to reduce the bias of the algorithm whereas bagging or Test-Time Augmentation decrease its variance.

Most ensembling methods increase the computing time (training time and/or prediction time) but provide an efficient way to push forward the state-of-the-art.
This study aims at reviewing the most relevant ensembling methods to be used for object detection based on VHR satellite imagery. The selected methods have been implemented in a Python-Keras framework and the performance gain of a combination of 3 methods is detailed in the last section.
%
%
%%%%%%%%%%%%%%%%%%%%%% State of the art
\section{State of the art of Ensembling techniques}
\subsection{Introduction}
All ensemble learning methods relies on combinations of variations in the following spaces:
\begin{enumerate}
\item Training data space: transformations of the data used to train the models
\item Model space: transformations of the models trained on the dataset
\item Prediction space: transformations and combinations of the prediction
\end{enumerate}
While research articles usually focus on single approaches in one of those categories, challenge-winning solutions use a combination of methods to achieve the highest performances.
In the following section, the ensemble learning methods relevant to object detection based on VHR satellite imagery will be listed according to the 3 categories previously introduced.

\begin{figure}[t!]
    \centering
    \includegraphics[width=12cm]{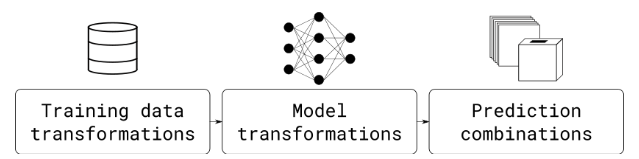}
    \caption{\label{fig:ensemble_pipeline}%
    \textbf{Ensembling pipeline architecture}}
\end{figure}

%%%%%%%%%%%%%%%%%%%%% Training data variation-based methods
\subsection{Training data variation-based methods}
\subsubsection{Bagging \cite{bagging_1} \cite{bagging_2}:} Bagging, or bootstrap aggregating, generates new data-sets by extracting random sub-samples from the original one. Training the model on those random sub-samples reduces the variance of the training data and improves the accuracy of final ensembling model. Compared to boosting, which samples the original dataset in order to correct the errors of the previous models iteratively, bagging can be used to train models in parallel. Another way of performing bagging without splitting the dataset, known as “synthetic bagging”, is to use data augmentation : models are then trained on the same dataset with different kinds of augmentations.

\begin{figure}[t!]
    \centering
    \includegraphics[width=12cm]{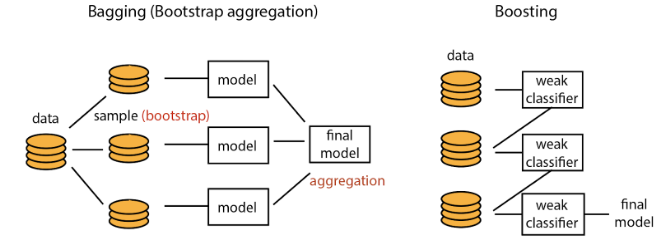}
    \caption{\label{fig:bagging}%
    \textbf{Comparison between Bagging and Boosting.}\newline
    \textbf{Details:} The main dataset is divided into 3 subsets that are used to train 3 replicas of a model. This is done in parallel for Bagging and iteratively for Boosting.}
\end{figure}
%
%
%%%%%%%%%%%%%%%%%%%%% Model variation-based methods
\subsection{Model variation-based methods}
In remote sensing, different CNN architectures are known to have different strong-points: some models might be better at separating adjacent objects, detecting smaller elements or providing more precise segmentation. Varying hyper-parameters, such as loss and optimizer, also provides interesting models to combine.
\subsubsection{Snapshot Ensembling methods:}

The eponym sub-method Snapshot Ensembling (SSE) \cite{sse} is a method providing ’cheaply’ a large number of complementary CNN that could be combined to improve performances. While those CNN share the same architecture, the use of a clever ``cyclical scheduling" of the learning rate allows the networks to reach different wide or narrow local optima along training. 
The learning rate is gradually decreased to allow the network to converge to some local minima. The algorithm then takes a snapshot of the weights, before increasing the learning rate again in order to ’jump’ to another optimum. This cycle can be repeated many times to produce a set of snapshots of the desired size.

Fast Geometric Ensembling (FGE) \cite{fge}  is a variation of snapshot ensembling using a different scheduling scheme: rapidly cycling linear piecewise variations of the learning rate instead of the original smooth cosine one. This approach is justified by the existence, according to the authors of \cite{fge}, of low-loss paths between interesting minimas. FGE is reportedly faster to train than SSE while showing performances improvements.

\begin{figure}[t!]
    \centering
    \includegraphics[width=12cm]{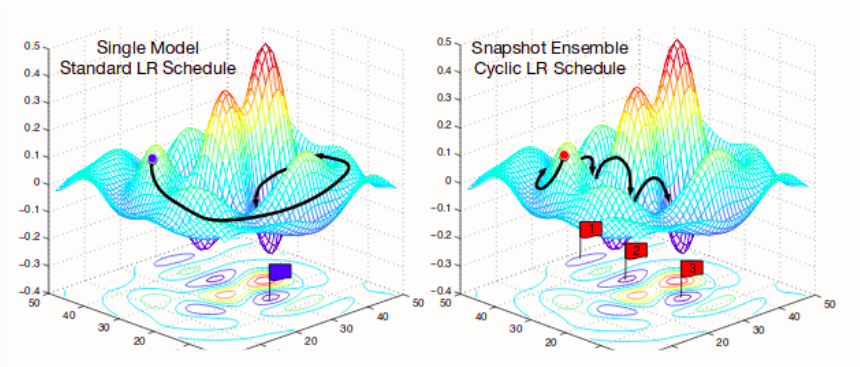}
    \caption{\label{fig:sse_loss}%
    \textbf{classical SGD optimization vs Snapshot Ensembling.}
    source: \cite{sse}}
\end{figure}

\subsubsection{Stochastic Weight Averaging (SWA) \cite{swa_1} \cite{swa_2}:}
Stochastic Weight Averaging is an optimization method approximating FGE while reducing the computational cost at prediction time. During training, a combination of two models: the first keeps a running average of the weights at each iteration, while the second model explores the weight space following a cycling learning rate scheduler. At inference, Stochastic Weight Averaging provides a single model with averaged weights.

\begin{figure}[t!]
    \centering
    \includegraphics[width=12cm]{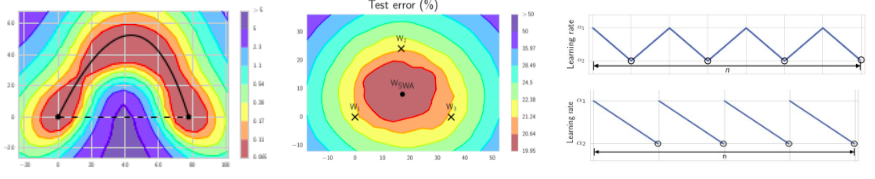}
    \caption{\label{fig:fge_swa}%
    \textbf{Illustration of FGE and SWA methods.}
    \textbf{Left:} example of a low error rate path between 2 local minimas (FGE method).
    \textbf{Middle:} example of 3 weights networks $W_{1}$, $W_{2}$ and $W_{3}$, and their weights average $W_{SWA}$ that is on a wider local minima with lower error rate (SWA method).
    \textbf{Right:} the corresponding scheduling schemes of the learning rate (FGE on the top, SWA on the bottom).
    Sources: \cite{fge} and \cite{swa_1}}.
\end{figure}

\subsubsection{Hydra \cite{hydra}:}
Hydra is a CNN ensembling method that consists of training an initial ``coarse" model, called the "body": a snapshot of the training with average performances. From this body, a set of "heads" are fined-tuned by applying various data augmentations. Just like snapshot ensembling, this approach provides a set of local minima to combine from a single neural network model.

\begin{figure}[t!]
    \centering
    \includegraphics[width=6cm]{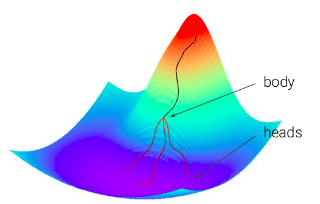}
    \caption{\label{fig:hydra}%
    \textbf{Illustration of the optimization process in the Hydra framework.} Source: \cite{hydra}}.
\end{figure}

%%%%%%%%%%%%%%%%%%%%% Prediction variation-based methods
\subsection{Prediction variation-based methods}

\subsubsection{Voting-based methods \cite{vote}:}
This very simple fusion method combines predictions by counting the agreeing detectors. This vote could be counted either on the raw pixel outputs of the detector or object-wised from the vectorised shapefile.

\subsubsection{Test Time Augmentation (TTA) \cite{tta}:}
CNN are not intrinsically invariant to translation, rotation, scale and other transformations. Applying such transformations to the input image before feeding it to the network can correct this variability and increase the performances of the model. While TTA does not need any new training, it increases linearly the prediction time.

\subsubsection{Stacked generalization \cite{stacking_1} \cite{stacking_2}:}
Stacking relies on a "meta-learner", which take the predictions of the individual detectors as its input and is trained to output a refined output. While stacking has the potential of learning the optimal fusion of detectors, It is very specific to the input detectors and is quite a rigid method as the "meta-learner" needs new to be re-trained if an input detector is modified.

\subsubsection{Dropout as a Bayesian Approximation \cite{dropout}:}
Dropout is the process of switching off random neurons within the network during training. During training, this method helps limiting the overfitting and can be seen as an ensemble of network sub-sampled in the model space. [14] introduces the idea of using dropout not only during training but also in prediction stage: several predictions are run by removing different random neurons from the network. Those predictions can then be averaged to achieve better generalisation.

%%%%%%%%%%%%%%%%%%%%%% Application
\section{Application} In this article, we present the results of the tests of three ensembling techniques on the vehicle detection tasks in VHR images.

\subsubsection{Use case:} 
\begin{figure}[t!]
    \centering
    \includegraphics[width=10cm]{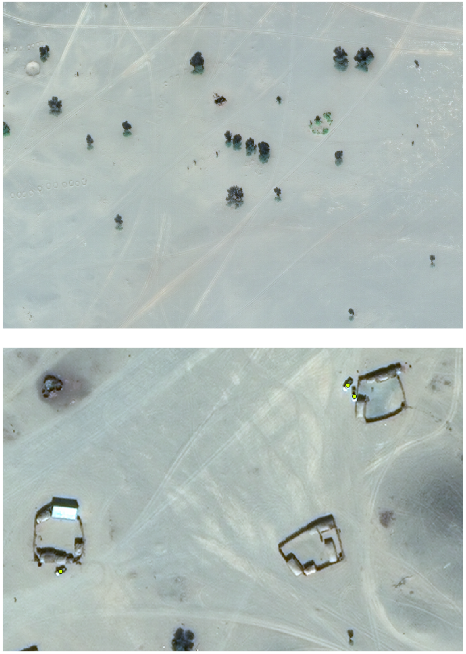}
    \caption{\label{fig:desert_example}%
    \textbf{Hard examples in the desert environment testing set where many objects can be confused with vehicles.}
    \textbf{Top:} example of small trees that look similar to vehicles.
    \textbf{Bottom:} example of very small buildings that look similar to vehicles (yellow points are the vehicles in the ground truth).}
\end{figure}

\begin{figure}[t!]
    \centering
    \includegraphics[width=12cm]{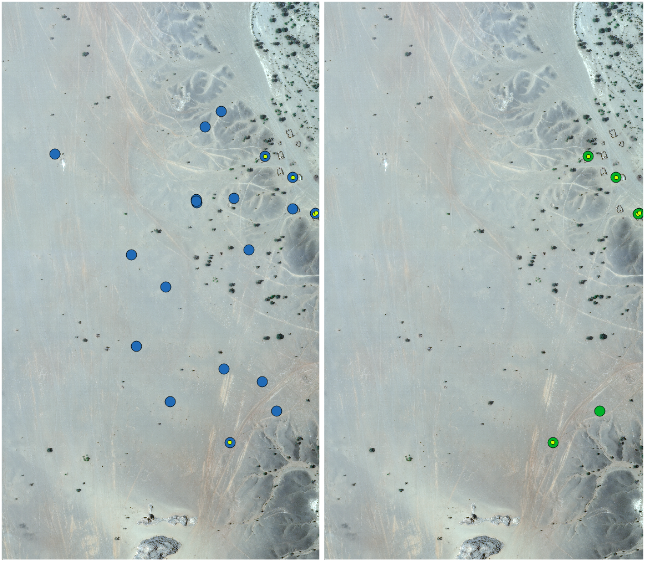}
    \caption{\label{fig:results_im_2}%
    \textbf{Example of detections on the desert environment testing set.}
    \textbf{Left:} Detection of the original model in blue (operating point 1) and ground truth in yellow.
    \textbf{Right:} Detection of the model with TTA and Bayesian Dropout combined in green (operating point 1) and ground truth in yellow.
    \textbf{Interpretation:} We observe on the left side that the original model achieves a very good recall with some false detections. The detection using ensembling removes most of these false detections without affecting the recall and increase the overall performance.}
\end{figure}

In object detection for military purposes, a use case of interest is detecting vehicles in the desert. On the operational side, it is very difficult and time consuming to manually find those small objects in the middle of very wide areas, as it can be seen on Fig. \ref{fig:results_im_2}. On the machine learning side, the task of detecting vehicles in such environment is very challenging, even for high performance detectors, since a lot of objects look like vehicles and are very difficult to distinguish (small trees and rocks, very small buildings, etc.), which leads to an increase of false positives. An example of such objects is shown on fig. \ref{fig:desert_example}.

\subsubsection{Original model:} We performed the tests of the ensembling techniques applied on a very generic and high performance model owned by Earthcube (92\% of F1-score on a diverse environment testing set of size 31.2 $km^2$ detailed in fig. \ref{fig:testing_sets}). It is a custom ResNet-Unet architecture model with 8.5M parameters. It has been trained for the segmentation task on a very large training dataset (330K civil vehicles labeled at a pixel scale) comprising VHR satellite images of different types of environments (urban, vegetal, desertic, coastal) and conditions (aerosol, summer, snow, etc.) from sites located all around the world.

\subsubsection{Testing set:} 
\begin{figure}[t!]
    \centering
    \includegraphics[width=12cm]{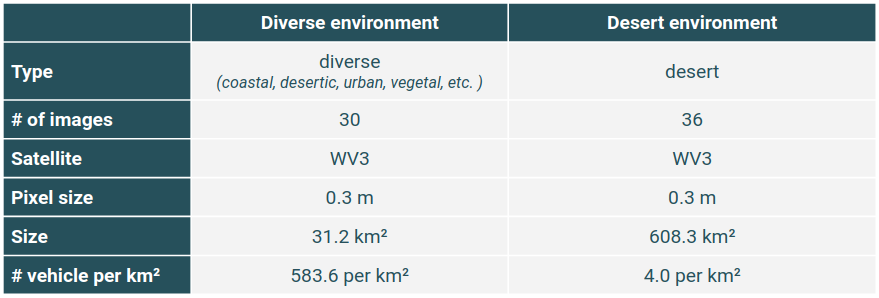}
    \caption{\label{fig:testing_sets}%
    \textbf{Information on the testing sets.}
    \textbf{Left:} Information on the diverse environment testing set (original model has 92\% of F1-score on this testing set).
    \textbf{Right:} Information on the desert environment testing set (very challenging for vehicle detection since objects look similar to vehicles).}
\end{figure}

In order to assess and compare the tested ensembling techniques on the use case, we used a very challenging testing set composed by images of desert environment that is particularly difficult for vehicle detection (detailed in fig. \ref{fig:testing_sets}). We used a very large testing set (composed of 36 different sites for a total area of 608.3 $km^2$) in order to make the assessment of the ensembling techniques relevant.

\subsubsection{Tested ensembling techniques:} Based on the state of the art survey, the ensembling techniques that are suitable to the considered use case have been selected. Since the training dataset is very large, the additional number of epochs to train the model on is a key constraint.

Three ensembling techniques have been tested: TTA combined with Bayesian Dropout, FGE and SWA. TTA and Bayesian Dropout do not need any additional training since are applied at prediction time. Based on previous experiments, the performance is increased when these two techniques are combined. On the other hand, FGE and SWA are promising and only need fine-tuning of the original model so they are not time-consuming to implement.

Other ensembling techniques that need a large number of epochs to train, such as Bagging, SSE, and Hydra, have been discarded (and may need a full article by themselves). Bagging is both time consuming and complex to train since the training dataset size is modified so new tests have to be performed to optimize the model size for each sub-datasets. SSE needs to train the model from scratch several times in a row and provides lower performance increase than FGE and SWA according to preliminary study. Finally, Hydra needs to ensure convergence of very different models to diversify augmentations and vary fundamental parameters which is both complex and time-consuming to tune and monitor.

\begin{figure}[t!]
    \centering
    \includegraphics[width=12cm]{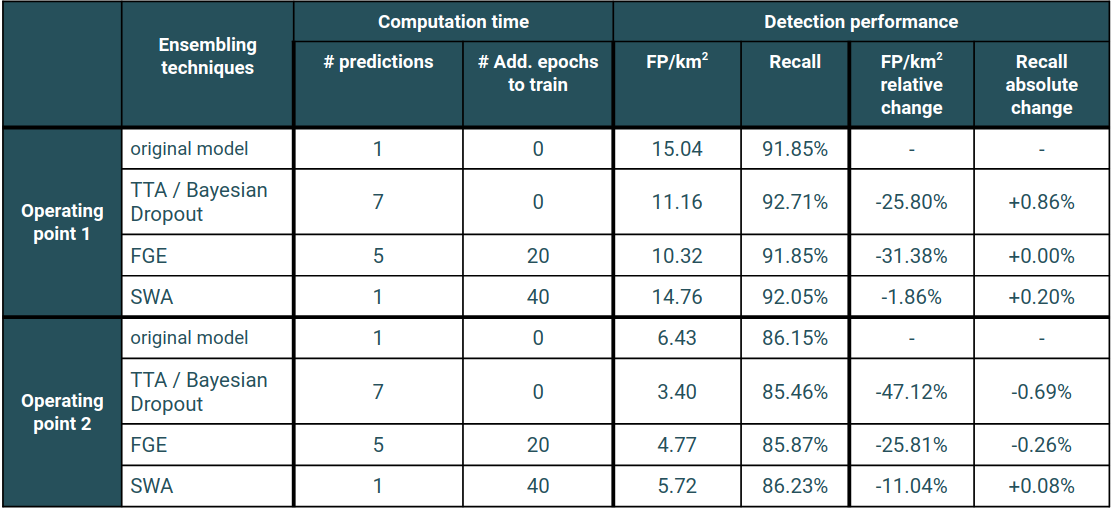}
    \caption{\label{fig:results_table}%
    \textbf{Results of the tests on ensembling techniques on the desert environment testing set.}
    \textbf{Interpretation:} Those techniques are compared on two operating points. The first one has a very high recall close to 92\%, the second one has a lower recall close to 86\% with less false detections. For each technique and each operating point, we assessed both the detection performance and the computation time.
    \textbf{Detection performance metric:} For the detection performance, we assessed the performance gain by using both the absolute increase in recall, and the relative decrease of false positives per $km^2$. We have not used the F1-score metric since it is not suitable for desert areas: there are very few vehicles in very large areas, which make the F1-score very unstable (see fig. \ref{fig:testing_sets}).
    \textbf{Computation time metric:} To assess the prediction time, we use the number of predictions to perform so that it does not depend on the available hardware or the image size. For training time, we used the number of additional epochs to train the model.
    \textbf{Parameters optimization:} The parameters of each technique (number of snapshots to fuse for FGE and SWA, and number and nature of the transformations for TTA / Bayesian Dropout) have been previously optimized. For example, a ratio of gain of performance versus additional computation time has been maximized.}
\end{figure}

\subsubsection{Result of the tests:} The results of the tests are gathered on the table in fig. \ref{fig:results_table}. The selected ensembling techniques have been tested on two high recall operating points.

TTA combined with Bayesian Dropout achieves a very large performance increase on such challenging desert environment: +0.86\% of recall while decreasing the number of false positives by 25\% for an operating point with  around 92\% of recall. This result means that this technique has operational value since it can boost very significantly on challenging environment a detector that has already operational performance.

FGE achieves a very similar performance boost to TTA / Bayesian Dropout: -31\% of false positives while achieving the same recall for the 92\% recall operating point. So it provides similar operational value than TTA / Bayesian.

Finally, SWA achieves a significant performance boost even though it is smaller than TTA / Bayesian Dropout and FGE: -11\% of false positives on the 85\% recall operating point, and small performance increase on the 92\% recall operating point.
However, this technique only needs one prediction (rather than respectively 7 and 5 for TTA / Bayesian Dropout and FGE). So this technique is highly recommended for applications that cannot use additional time during prediction. One could conclude that for a very high level of recall (+90\%), substantial performance boosts can only be achieved through voting-based techniques (such as TTA / Bayesian and FGE) and that using one unique model might not be sufficient.

\begin{figure}[t!]
    \centering
    \includegraphics[width=12cm]{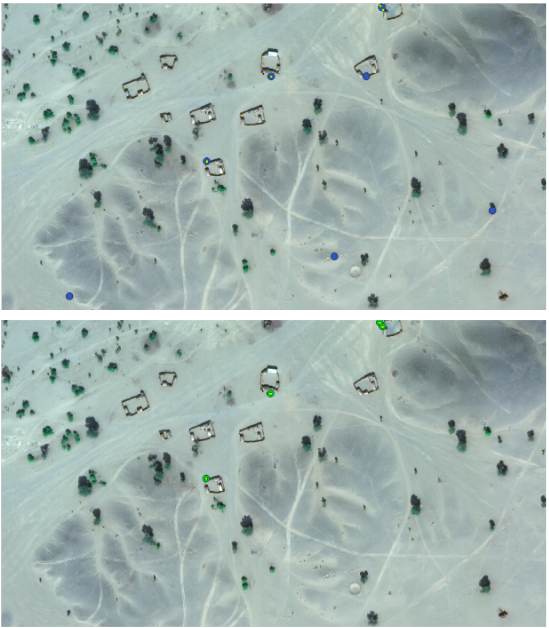}
    \caption{\label{fig:results_im_1}%
    \textbf{Example of detections on the desert environment testing set.}
    \textbf{Top:} Detection of the original model in blue (operating point 1) and ground truth in yellow.
    \textbf{Bottom:} Detection of the model with TTA and Bayesian Dropout combined in green (operating point 1) and ground truth in yellow.
    \textbf{Interpretation:} We observe that the original model has a high recall and provides some false detections on small trees and buildings that look similar to vehicles. The combination of TTA with Bayesian Dropout removes these false dections and increase the recall by detecting an additional vehicle on the top.}
\end{figure}

%%%%%%%%%%%%%%%%%%%%%% Conclusion
\section{Conclusion}
Ensembling techniques provide a toolbox to increase performances of a given model with a given training dataset, at each level of the detection pipeline (training data, model, prediction).

After detailing the most relevant ensembling techniques for object detection in VHR images, three techniques have been selected for the use case of vehicle detection on VHR images with an original very high performance model trained on a very large dataset. These techniques have been assessed and compared and provide substantial performance increases on a very challenging desert environment testing set.

At the price of extra training/prediction time, ensembling levels up state of the art algorithms and brings automatic vehicle detection at the level of a human expert. In general, the work presented here allows for better performances and higher reliability for any object detection solution. This can give the armed forces better tools to perform their missions, enabling them to automatically extract relevant information from satellite images.

This can be used for specific tasks such as identifying isolated vehicles in desertic areas or measuring activity at a given location but also for more broadly defined tasks such as strategic sites monitoring. It can be used for any type of object (vehicles, vessels, buildings, roads, etc.) and can either be deployed alone without any human looking at the images or be applied as a tool to help image analysts in their missions.


\end{document}